# Improving Heuristics Through Relaxed Search
## – An Analysis of TP4 and $\text{HSP}^*_\text{a}$ in the 2004 Planning Competition

**Patrik Haslum**                                                    PAHAS@IDA.LIU.SE
*Linköpings Universitet*
*Linköping, Sweden*

## Abstract

The $h^m$ admissible heuristics for (sequential and temporal) regression planning are defined by a parameterized relaxation of the optimal cost function in the regression search space, where the parameter $m$ offers a trade-off between the accuracy and computational cost of the heuristic. Existing methods for computing the $h^m$ heuristic require time exponential in $m$, limiting them to small values ($m \leqslant 2$). The $h^m$ heuristic can also be viewed as the optimal cost function in a relaxation of the search space: this paper presents *relaxed search*, a method for computing this function partially by searching in the relaxed space. The relaxed search method, because it computes $h^m$ only partially, is computationally cheaper and therefore usable for higher values of $m$. The (complete) $h^2$ heuristic is combined with partial $h^m$ heuristics, for $m = 3, \ldots$, computed by relaxed search, resulting in a more accurate heuristic.

This use of the relaxed search method to improve on the $h^2$ heuristic is evaluated by comparing two optimal temporal planners: TP4, which does not use it, and $\text{HSP}^*_\text{a}$, which uses it but is otherwise identical to TP4. The comparison is made on the domains used in the 2004 International Planning Competition, in which both planners participated. Relaxed search is found to be cost effective in some of these domains, but not all. Analysis reveals a characterization of the domains in which relaxed search can be expected to be cost effective, in terms of two measures on the original and relaxed search spaces. In the domains where relaxed search is cost effective, expanding small states is computationally cheaper than expanding large states and small states tend to have small successor states.

## 1. Introduction

In the 2004 International Planning Competition, I entered two planners: TP4 and $\text{HSP}^*_\text{a}$. TP4, which also participated in the 2002 competition, is a temporal STRIPS planner, optimal *w.r.t.* makespan. It is based on temporal regression search with an admissible heuristic called $h^2$. The temporal $h^2$ heuristic is an instance of the general $h^m$ ($m = 1, 2, \ldots$) family of heuristics, which is defined by a parameterized relaxation of the optimal cost function over the search space. $\text{HSP}^*_\text{a}$ is identical to TP4 except that it uses a recently developed method, called *relaxed search*[1], to improve the $h^2$ heuristic (Haslum, 2004a).

Regression planners carry out a search over sets of goals, starting from the goal given in the planning problem. The relaxation that leads to the $h^m$ heuristics is to assume that the cost of any set of more than $m$ goals equals the cost of the most costly subset of size $m$. The heuristic can underestimate the cost of a goal set, if there are interactions involving more than $m$ goals, but it can never overestimate. In the TP4 and $\text{HSP}_\text{a}^*$ temporal planners, the cost of a set of goals is the minimum total execution time, or makespan, of any plan that achieves the goals, but the same relaxation can be used to derive heuristics for regression planning with different plan structures and cost measures, $e.g.$, sequential plans with sum of action costs (Haslum, Bonet, & Geffner, 2005), or to estimate resource consumption (Haslum & Geffner, 2001). Formally, the $h^m$ heuristic, for any $m$, can be defined as the solution to a relaxation of the optimal cost equation (known as the Bellman equation) which characterizes the optimal cost function over the search space. A complete solution to the relaxed equation is computed explicitly, by solving a generalized shortest path problem, prior to search and stored in a table which is used to calculate heuristic values of states during search.

---

1. Haslum (2004a) called the method "approximate search", an unfortunate choice of name as it evokes the wrong connotations. The term relaxed search, which better corresponds to the intuition underlying the method, will be used in this paper.

The parameter $m$ offers a trade-off between the accuracy of the heuristic and its computational cost: the higher $m$, the more subgoal interactions are taken into account and the closer the heuristic is to the true cost of all goals in a state, while on the other hand, computing the solution to the relaxed cost equation is polynomial in the size of the problem (the number of atoms) but exponential in $m$. Because the current method for computing the heuristic computes a $complete$ solution to the relaxed cost equation, the heuristic exhibits for most planning problems a "diminishing marginal gain": once $m$ goes over a certain threshold (typically, $m = 2$) the improvement brought by the use of $h^{m+1}$ over $h^m$ becomes smaller for increasing $m$. This combines to make the method cost effective, in the sense that the heuristic reduces search time more than the time required to compute it, only for small values of $m$ (typically, $m \leqslant 2$). However, the $h^2$ heuristic is often too weak. The question addressed here is if a more accurate – and cost effective – heuristic can be derived in the $h^m$ framework. The idea of relaxed search is to compute $h^m$ (for higher $m$) only $partially$ to avoid the exponential increase in computational cost. The alternative would of course be to abandon the $h^m$ framework and look at other approaches to deriving admissible heuristics for optimal temporal planning, but there are not many to be found: existing makespan-optimal temporal planners either use the temporal $h^2$ heuristic ($e.g.$, CPT, Vidal & Geffner, 2004) or obtain estimates from a temporal planning graph ($e.g.$, TGP, Smith & Weld, 1999, and TPSys, Garrido, Onaindia, & Barber, 2001), which also encodes the $h^2$ heuristic though computed in a different fashion. The domain-independent heuristics used in other temporal planners, such as LPGP (Long & Fox, 2003) or IxTeT (Trinquart, 2003), are used to estimate the distance to the nearest solution in the search space, rather than the cost ($i.e.$, makespan) of that solution.

The relaxation underlying the $h^m$ heuristics can be explained in terms of the search space, rather than in terms of solution cost: any set of more than $m$ goals is "split" into problems of $m$ goals each, which are solved independently (the split is not a partitioning, since $all$ subsets of size $m$ are solved). The relaxed cost equation is also the optimal cost equation for this relaxed search space, which I'll call the $m$-$regression$ space. The relaxed search method consists in solving the planning problem ($i.e.$, searching from the top level goals) in the $m$-regression space. During the search, parts of the solution to the relaxed cost equation are discovered, and these are stored in a table for later use, just as in the previous approach. Because the relaxed search only visits $part$ of the $m$-regression search space, it can be expected to be able to do so more quickly than methods that build a solution to the cost equation for the entire $m$-regression space. Consequently, it can be applied for higher values of $m$. Since the relaxed search is done from the goals of the planning problem, the part of the $h^m$ solution that is computed is also likely be the most relevant part. The complete and partial $h^m$ heuristics computed for different $m$ by the two methods can be combined by maximization, resulting in a more accurate final heuristic, hopefully at a computational cost that is not greater than its value.

This paper makes two main contributions: First, it provides a detailed presentation of the relaxed search method and how this method is applied to classical and temporal regression planning. Although the method is presented in the context of planning, it is quite general and may be applicable to other search problems as well. Indeed, similar techniques have been applied to planning and other single agent search problems (Prieditis, 1993; Junghanns & Schaeffer, 2001) and to constraint optimization (Ginsberg & Harvey, 1992; Verfaillie, Lemaitre, & Schiex, 1996). The relation to these ideas and techniques is also discussed. Second, it presents the results of an extended analysis of the relative performance of TP4 and $\text{HSP}_a^*$ in the domains of the planning competition. The picture that emerges from this analysis is somewhat different from that given by the competition results. In part this is due to the time-per-problem limit imposed in the competition, since the advantage of $\text{HSP}_a^*$ over TP4 is mainly on "hard" problems, which require a lot of time to solve for both planners. In part, it is also because the version of $\text{HSP}_a^*$ used in the competition was buggy. The main result of the analysis, however, is a characterization of the domains in which relaxed search can be expected to be cost effective: in such domains, expanding small states is computationally cheaper than expanding large states, and small states tend to have small successor states. It is also shown that these criteria can be (weakly) quantified by two measures, involving the relative regression branching factors and the size of states generated by regression, in the original and relaxed search spaces.

## 2. Background

The TP4 planner finds temporal plans for STRIPS problems with durative actions. The plans found are optimal *w.r.t.* makespan, *i.e.*, the total execution time of the plan[2], and the planner is also able to ensure that plans do not violate certain kinds of resource constraints.

The main working principles of TP4 are a formulation of a regression search space for temporal planning and the $h^m$ family of admissible heuristics, brought together through the IDA* search algorithm. These, and the overall architecture of the planner, are briefly described in this section; more details on the planner can be found in earlier papers (Haslum & Geffner, 2000, 2001; Haslum, 2004b). To provide background for a clearer description of relaxed search in the next section, search space and heuristics are explained first for the simpler case of sequential planning, followed by their adaption to the temporal case. Also, certain technical details that appear important in explaining the behaviour of $\text{HSP}_a^*$ relative to TP4 in the competition domains will be highlighted.

---

2. It should be noted, however, that the plan makespan is optimal with respect to the semantics that TP4 assumes for temporal planning, which differs somewhat from that specified for PDDL2.1 (see Section 2.2.1). To make plans acceptable to the PDDL2.1 plan validator it is necessary to insert some "whitespace" into the plan, increasing the makespan slightly.

## 2.1 Regression Planning: Sequential Case

We assume the standard propositional STRIPS model of planning. A planning problem ($P$) consists of a set of atoms, a set of actions and two subsets of atoms: those true in the initial state ($I$) and those required to be true in the goal state ($G$). Each action $a$ is described by a set of precondition atoms ($pre(a)$), which have to hold in a state for the action to be executable, and sets of atoms made true ($add(a)$) and false ($del(a)$) by the action. A solution plan is an executable sequence or schedule of actions ending in a state where all goal atoms hold. The exact plan form depends on the measure optimized: in the sequential case, a cost is associated to each action ($cost(a) > 0$), a plan is a sequence of actions, and the sum of their costs is the cost of the plan.

Regression is a planning method in which the search for a plan is made in the space of "plan tails", partial plans that achieve the goals provided that the preconditions of the partial plan are met. Search ends when a plan tail whose preconditions are already satisfied by the initial state is found. For sequential planning, the preconditions provide a sufficient summary of the plan tail. Thus, a sequential regression state is a set, $s$, of atoms, representing subgoals to be achieved. An action $a$ can be used to regress a state $s$ iff $del(a) \cap s = \emptyset$, and the result of regressing $s$ through $a$ is $s' = (s - add(a)) \cup pre(a)$. The search starts from the set of goals $G$ and ends when a state $s \subseteq I$ is reached.

## 2.2 Regression Planning: Temporal Case

In the case of temporal planning, each action has a duration ($dur(a) > 0$). The plan is a schedule, where actions may execute in parallel (subject to resource and compatibility constraints), and the objective to minimize is the total execution time, or makespan. When action durations are all equal to 1, the special case of parallel planning results.

### 2.2.1 A Note on PDDL2.2 Compliance

TP4 and HSP$_a^*$ do not support any of the new features introduced in PDDL2.2, the problem specification language for the 2004 competition (Edelkamp & Hoffmann, 2004). The planners support durative actions, obviously, but these are interpreted in a manner that differs from the PDDL2.1 specification (Fox & Long, 2003). For practical purposes, TP4 and HSP$_a^*$ accept the PDDL2.1 syntax. Numeric state variables (called "fluents" in PDDL2.1) are supported only in certain forms of use.

The semantics that TP4 and HSP$_a^*$ assume for durative actions are essentially those introduced by Smith and Weld (1999) for the TGP planner. For an action $a$ to be executable over a time interval $[t, t + dur(a)]$, atoms in $pre(a)$ must be true at $t$, and persistent preconditions (atoms in $per(a) = pre(a) - del(a)$) must remain true over the entire interval. Effects of the action take place at some point in the interior of the interval, and can be relied on to hold at the end point. Two actions, $a$ and $a'$, are assumed to be *compatible*, in the sense that they can be executed in overlapping intervals without interfering with each other iff neither action deletes an atom that is a precondition of or added by the other, *i.e.*, iff $del(a) \cap pre(a') = del(a) \cap add(a') = \emptyset$ and vice versa.

This interpretation of durative actions respects the "no moving target" rule of PDDL2.1, but in a different way: instead of requiring plans to explicitly separate an action depending on a condition from the effect that establishes the condition, the semantics requires that

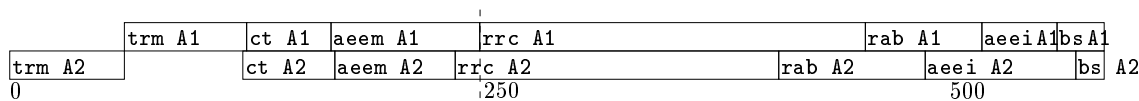

Figure 1: A temporal plan (the solution to problem `p06` from the `umts` domain). The plan also contains two actions (`am A1`) and (`am A2`), which are not visible because they have zero duration. Actions (`trm A1`) and (`trm A2`) are separated because of a resource conflict. The makespan of the plan is 582.

*change takes place in a time interval.* This makes durative actions strictly less expressive than in PDDL2.1, where effects can be specified to take place exactly at the start or end of an action. In particular, it does not support actions that make a condition true only during their execution (*i.e.*, add an atom at the start of the action and delete it again at the end), which prevented TP4 and HSP$_a^*$ from solving the compiled versions of problems with timed initial literals.

In principle, it is certainly possible to devise a temporal regression search space for the PDDL2.1 interpretation of durative actions, although states in this space would be far more complex structures, due to the need to retain more of the plan tail in the state (enough to include the end point of all on-going actions). The LPGP (Long & Fox, 2003) and TPSys (Garrido, Fox, & Long, 2002) planners both use the PDDL2.1 semantics, and are both Graphplan derivatives and thus carry out a search resembling regression in their solution extraction phase (though both planners embody modifications to the purely back-chaining solution extraction used in Graphplan). However, in the planning domains that have been used in the two planning competitions since the introduction of temporal planning into PDDL, and also in most of the example domains that have appeared in the literature, the main use of the stronger PDDL2.1 semantics of durative actions has been to encode certain features, such as the timed initial literals used in some domain versions in the last competition, or "non-inert" facts (*i.e.*, facts that do not persist over time unless maintained by an action). It may very well be easier to add some of these features directly to the temporal regression formulation used by the TP4 and HSP$_a^*$ planners, though this has yet to be put to the test.

Numeric state variables that are used (by actions) in certain specific ways are interpreted as resources (or cost measures, in sequential planning) and supported by the planners, though with some restrictions. The unrestricted use of numeric state variables allowed by PDDL2.1 is not supported. A more detailed discussion can be found in the paper on TP4 and HSP$_a^*$ in the competition booklet (Haslum, 2004b).

### 2.2.2 TEMPORAL REGRESSION

Temporal regression, just like sequential regression, is a search in the space of plan tails. However, in the temporal case the set of precondition atoms is no longer sufficient to summarize a plan tail: states have to be be extended with actions concurrent with the subgoals and the timing of those actions relative to the subgoals. Consider the example plan in Figure 1, specifically the "state" at time 250: since this is the starting point of action (`rrc`

`A1`), its preconditions must be goals to be achieved at this point. But the actions (including no-ops) establishing those conditions must be compatible with the action (`rrc A2`), which starts 13 units of time earlier and whose execution spans across this point.

Thus, a temporal regression search state is a pair $s = (E, F)$, where $E$ is a set of atoms and $F = \{(a_1, \delta_1), \ldots, (a_n, \delta_n)\}$ is a set of actions $a_i$ with time increments $\delta_i$. This represents a partial plan (tail) where the atoms in $E$ must hold and each action $(a_i, \delta_i)$ in $F$ has been started $\delta_i$ time units earlier. Put another way, an executable plan (schedule) *achieves* state $s = (E, F)$ at time $t$ iff the plan makes all atoms in $E$ true at $t$ and schedules action $a_i$ at time $t - \delta_i$ for each $(a_i, \delta_i) \in F$.

When expanding a state $s = (E, F)$, successor states $s' = (E', F')$ are constructed by choosing (non-deterministically) for each atom $p \in E$ an establisher (*i.e.*, a regular action or no-op $a$ with $p \in add(a)$), such that chosen actions are compatible (as defined in Section 2.2.1) with each other and with all actions in $F$, and advancing time to the next point where an action starts (since this is a regression search, "advancing" and "next" are in the direction of the beginning of the developing plan). Preconditions of all actions and no-ops starting at this point become $E'$ while remaining actions (with their time increments adjusted) become $F'$. A state $s = (E, F)$ is final if $F = \emptyset$ and $E \subseteq I$.

The exact details of the temporal regression search are not important for the rest of this paper and have been described elsewhere (Haslum & Geffner, 2001).

### 2.2.3 Right-Shift Cuts

In a temporal plan there is usually some "slack", *i.e.*, some actions can be shifted forward or backward in time without changing the structure or makespan of the plan. A right-shifted plan is one in which all such movable actions are scheduled as late as possible. Non-right-shifted plans can be excluded from consideration without endangering optimality. Doing this eliminates redundant branches in the search space, which often speeds up planning significantly[3].

This can be achieved by applying the following rule: When expanding a state $s' = (E', F')$ with predecessor $s = (E, F)$, an action $a$ compatible with all actions in $F$ may *not* be used to establish an atom in $s'$ when all the atoms in $E'$ that $a$ adds have been obtained from $s$ by no-ops. The reason is that $a$ could have been used to support the same atoms in $E$, and thus could have been shifted to the right (delayed).

Again, details can be found elsewhere and are not important. What *is* important to note is that the right-shifting rule refers to the predecessor of the state being expanded. This means that when the rule is applied, the possible successors to, and therefore the optimal cost of, a regression state may be different depending on the path through which the state was reached. Thus, the lower bound on the cost of a state obtained when the state is expanded but not solved (as it will be in an IDA* search) may be invalid as a lower bound for the same state when reached via a different path.

---

3. From an execution point of view, it may be preferable to place actions whose execution time in the plan is not precisely constrained as early as possible (*i.e.*, left-shifted) rather than at the latest possible time. From a search point of view what matters is that of the many possible, but structurally equivalent, positions in time for an action, only one is considered. The reason why right-shifting is used instead of left-shifting is that in a regression search, a left-shift rule will trigger later (*i.e.*, deeper in the search tree) and thus provide less efficient pruning.

## 2.3 Admissible Heuristics: Sequential Case

Let $h^*(s)$ denote the optimal cost function, *i.e.*, the function that assigns to each state $s$ in the search space the minimal cost of any path from $s$ to a final state (a state $s' \subseteq I$, in the regression planning space). The function $h^*(s)$ is characterized by the Bellman equation (Bellman, 1957):

$$h^*(s) = \begin{cases} 0 & \text{if } s \subseteq I \\ \min_{s' \in succ(s)} h^*(s') + \delta(s, s') \end{cases} \tag{1}$$

where $succ(s)$ is the set of successor states to $s$, *i.e.*, the set of states that can be constructed from $s$ by regression, and $\delta(s, s')$ is the "delta cost", *i.e.*, the increase in accumulated cost between $s$ and $s'$. In the sequential setting, this equals the cost of the action used to regress from $s$ to $s'$. Equation 1 characterizes $h^*(s)$ only on states $s$ that are reachable: the cost of an unreachable state is defined to be infinite.

Because achieving a regression state (*i.e.*, set of goals) $s$ implies achieving all atoms in $s$, and therefore any subset of $s$, the optimal cost function satisfies the inequality

$$h^*(s) \geqslant \max_{s' \subseteq s, |s'| \leqslant m} h^*(s') \tag{2}$$

for any $m$. Assuming that this inequality is actually an equality is the relaxation that gives the $h^m$ heuristics: rewriting equation (1) using (2) as an equality results in

$$h^m(s) = \begin{cases} 0 & \text{if } s \subseteq I \\ \min_{s' \in succ(s)} h^m(s') + \delta(s, s') & \text{if } |s| \leqslant m \\ \max_{s' \subseteq s, |s'| \leqslant m} h^m(s') \end{cases} \tag{3}$$

A complete solution to this equation, in the form of an explicit table of $h^m(s)$ for all sets with $|s| \leqslant m$, can be computed by solving a generalized single-source-all-targets shortest path problem. A variety of algorithms (all variations of dynamic programming or generalized shortest path) can be used to solve this problem, as described by, *e.g.*, Liu *et al.* (2002). TP4 and $\text{HSP}_a^*$ use a variation of the Generalized Bellman-Ford (GBF) algorithm. Computing a complete solution to equation (3) is polynomial in the number of atoms but exponential in $m$, simply because the number of subsets of size $m$ or less grows exponentially with $m$. This limits the complete solution approach to small values of $m$ (in practice, $m \leqslant 2$).

### 2.3.1 ON-LINE EVALUATION AND THE HEURISTIC TABLE

The solution to equation (3) is stored in a table (which will be referred to as the *heuristic table*). The stored solution, however, comprises only values of $h^m(s)$ for sets $s$ such that $|s| \leqslant m$. To obtain the heuristic value of an arbitrary state, the last clause of equation (3) is evaluated "on-line", and during this evaluation the value of $h^m(s')$ for any $s'$ such that $|s'| \leqslant m$ is obtained by looking it up in the table.

In fact, the heuristic table implemented in TP4 and $\text{HSP}_a^*$ is a general mapping from sets of atoms to their associated value, and the heuristic value of a state $s$ is the maximal value of any subset of $s$ that is stored in the table. In other words, if $T(s)$ denotes the value stored for $s$, the heuristic value of a state $s$ is given by $h(s) = \max \{T(s') \, | \, s' \subseteq s, T(s') \text{ exists}\}$. When all and only sets of size $m$ or less are stored in the table (as is the case when $h^m$ is computed completely) this coincides with evaluating the last clause of equation (3). However, the use of a general heuristic table implies that as soon as a value for *any* atom set $s$ is stored in the table, it becomes immediately included in all subsequent evaluations of states containing $s$. In particular, by storing parts of the solution to $h^{m'}$, for some higher $m'$, in the form of updates of the values of some size $m'$ atom sets, the heuristic evaluation implicitly computes the maximum of $h^m$ and the partially computed $h^{m'}$.

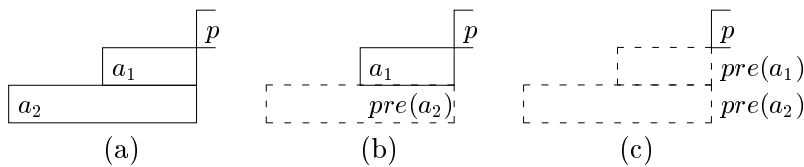

Figure 2: Relaxation of temporal regression states.

The heuristic table is implemented as a Trie (see *e.g.* Aho, Hopcroft, & Ullman, 1983) so that the evaluation of an atom set $s$ can be done in time linear in the number of subsets of $s$ that exist in the table[4]. Even so, there is some overhead compared to a table and evaluation procedure designed for a fixed maximal subset size.

## 2.4 Admissible Heuristics: Temporal Case

To define $h^m$ for temporal regression planning, one needs only to define a suitable measure of size for temporal regression states and then proceed as in the sequential case. Recall that a temporal regression state consists of two components, $s = (E, F)$, where $E$ is a set of atoms and $F$ a set of scheduled actions with time increments. The obvious candidate is to define $|s| = |E| + |F|$, and indeed, using this measure in equation (3) above results in a characterization of a lower bound function on the temporal regression space. In this case, however, due to the presence of a time increment $\delta$ in each $(\delta, a) \in F$, the set of states with $|s| \leqslant m$ is potentially infinite, and therefore the solution to this equation can not be computed explicitly.

To obtain a usable cost equation, a further relaxation is needed: since a plan that achieves the state $s = (E, F)$, for $F = \{(a_1, \delta_1), \ldots, (a_n, \delta_n)\}$, at time $t$ must achieve the preconditions of each action $a_i$ at time $t - \delta_i$, and these must remain true until $t$ unless

---

4. The claim of linear time evaluation holds only under the assumption of a certain regularity of entries stored in the table: The Trie data structure stores mappings indexed by strings, and the implementation of the heuristic table treats atom sets as strings in which atoms appear in a fixed lexical order. When an atom set $s$ is stored in the table, every set that is a prefix of $s$ viewed as a string in this way must also be stored, with value 0 if no better value is available. Due to the way heuristic values are computed (by complete solution of the $h^m$ equation or by relaxed search), this does not present a problem since whenever a set is stored, all its subsets (including the subsets corresponding to lexical prefixes) have already been stored. However, this is the reason why the heuristic table, in its current form, is not a substitute for the transposition table used in IDA* search.

deleted by $a_i$, the optimal cost function satisfies[5]

$$h^*(E, F) \quad \geqslant \quad \max_{(a_k, \delta_k) \in F} \left[ h^* \left( \bigcup_{(a_i, \delta_i) \in F, \, \delta_i \geqslant \delta_k} pre(a_i), \emptyset \right) + \delta_k \right] \tag{4}$$

$$h^*(E, F) \quad \geqslant \quad h^* \left( E \cup \bigcup_{(a_i, \delta_i) \in F} pre(a_i), \emptyset \right). \tag{5}$$

An example may clarify the principle: Consider the state $s = (\{p\}, \{(a_1, 1), (a_2, 2)\})$, depicted in Figure 2(a). A plan achieving this state at time $t$ must achieve the preconditions of $a_2$ at $t-2$, so $h^*(s)$ must be at least $h^*(pre(a_2), \emptyset) + 2$. If action $a_2$ is "left out", as in Figure 2(b), it can be seen that the same plan also achieves the joint preconditions of actions $a_1$ and $a_2$ at $t-1$, so $h^*(s)$ must be at least $h^*(pre(a_1) \cup pre(a_2), \emptyset) + 1$. Finally, if both actions are left out (Figure 2(c)), it is clear that the plan also achieves simultaneously the preconditions of the two actions and atom $p$, so $h^*(s)$ must be at least $h^*(\{p\} \cup pre(a_1) \cup pre(a_2), \emptyset)$.

By treating inequalities (4) – (5) as equalities, a temporal regression state is relaxed to a set of states in which $F = \emptyset$, *i.e.*, states containing only goals and no concurrent actions. To each such state, relaxation (2) can be applied, resulting in an equation defining temporal $h^m$, similar to (3). This equation has a finite explicit solution, containing all states $s = (E, \emptyset)$ with $|E| \leqslant m$. Again, more details can be found elsewhere (Haslum & Geffner, 2001).

## 2.5 IDA*

IDA* is a well known admissible heuristic search algorithm (see *e.g.* Korf, 1985, 1999). The algorithm works by a series of cost-bounded depth-first searches. The cost returned by the last completed depth-first search is a lower bound on the cost of any solution. Therefore, the algorithm can easily be modified to take an upper limit on solution cost, and to exit with failure once it has proven that no solution with a cost within this limit exists.

An extension of the IDA* algorithm for searching AND/OR graphs is the main tool by which the relaxed search method is implemented. The extended algorithm is presented in Section 3.2.

IDA* is a so-called linear space algorithm: it stores only the path to the current node. The algorithm can be speeded up by using memory in the form of a transposition table, which records updated estimated costs of nodes that have been expanded but not solved (Reinfeld & Marsland, 1994). The table is of a fixed limited size, so not all expanded unsolved nodes are stored[6]. Whenever the search reaches a node that is in the table the updated cost estimate for the node (discovered when the node was previously expanded) is used instead of its heuristic value, allowing the algorithm to avoid re-searching nodes that are reachable via several paths during the same iteration.

---

5. Note that one set of parentheses has been simplified away from $h^*(E, F)$: since a state $s = (E, F)$ is a pair, it should in fact be written "$h^*((E, F))$". Thus, the empty set in the right hand side of both inequalities is the second part of the state, *i.e.*, the set of concurrent actions $F$. This simplified form, with only a single pair of parentheses, is used throughout this paper.

6. This does *not* affect completeness or optimality of the search, since states that are not stored (due to collisions) are simply re-expanded if encountered again. In TP4 and HSP$^*_a$, the table is implemented as a closed hashtable, *i.e.*, states are stored only at the position corresponding to their hash values (thus lookup consists only in a single hash function computation, plus a state equality test to verify that the stored state is indeed the same as the one being looked up). In case of collisions, preference is given to storing nodes closer to the root of the search tree (Reinfeld & Marsland, 1994).

## 2.6 TP4

The TP4 planner precomputes the temporal $h^2$ heuristic as described above and uses it in an IDA* search in the temporal regression space. Right-shift cuts are used to eliminate redundant paths from the search space, and a transposition table is used to speed up search (Haslum & Geffner, 2001). The main steps of the planner are outlined in Figure 6 (on page 248), mainly to illustrate similarity and difference *w.r.t.* the $\text{HSP}^*_a$ planner.

# 3. Improving Heuristics Through Search

For many planning problems the $h^2$ heuristic is too weak. A more accurate heuristic can be obtained by considering higher values of the $m$ parameter, but any method for computing a complete solution to the $h^m$ equation scales exponentially in $m$, making it impractical for $m > 2$. A complete solution is useful because it helps detect unreachable states (in particular, $h^2$ detects a significant part of the static mutex relations in a planning problem), but also wasteful because often many of the atom sets are not relevant for evaluating states actually encountered while searching for a solution to the planning problem at hand. Recall that the heuristic evaluation of a state (a set of goals) makes use of the estimated cost of any subset of the state that is known (stored in the heuristic table). As larger atom sets are considered, *i.e.*, as $m$ increases, they become both more numerous and more specific, and thus the fraction of the complete solution that is actually useful decreases.

To use $h^m$ for higher $m$, clearly a way is needed to compute the heuristic at a cost proportionate to the value of the improvement. Relaxed search aims to achieve this by computing only a part of the $h^m$ solution, and a part that is likely to be relevant for solving the given planning problem.

## 3.1 Relaxed Search: Sequential Case

As explained earlier, the $h^m$ heuristic can be seen as the optimal cost function in the $m$-regression space, a relaxed search space where sets of more than $m$ goals are split into problems of $m$ goals, each of which is solved independently. Thus, the $m$-regression space is an AND/OR graph: states with $m$ or fewer atoms are OR-nodes and are expanded by normal regression, while states with more than $m$ atoms are AND-nodes, which are expanded by solving each subset of size $m$. The cost of an OR-node is minimized over all its successors, while the cost of an AND-node is maximized. Examples of (part of) this graph, for a 2-regression space, are shown in Figures 4 and 5 (the example is described in detail in Section 3.2.1). As can be seen, the graph is not strictly layered, in that OR-nodes may sometimes have successors that are also OR-nodes.

The different algorithms used to obtain complete solutions to the $h^m$ equation can all be seen as variations of a "bottom-up" labeling of the nodes of this graph, starting from nodes with cost zero and propagating costs to parent nodes according to this min/max principle. The propagation is complete, *i.e.*, proceeds until every (solvable) node in the graph has been labeled with its optimal cost (although in some of the algorithms, including the GBF implementation used by TP4 and $\text{HSP}_a^*$, only the costs of OR-nodes are actually stored). Relaxed search explores the $m$-regression space in a more focused fashion, with the aim of discovering the optimal cost (or an improved lower bound) of states relevant to the search for a solution to the goals of the given planning problem. This is achieved by searching the $m$-regression space for an optimal solution to a particular state: the cost of this solution is the $h^m$ heuristic value of that state. The algorithm described in the next section (IDAO*) carries out this search "top-down", starting from the state corresponding to the goals of the planning problem.

Heuristics derived by searching in an abstraction of the search space have been studied extensively in AI (see *e.g.* Gaschnig, 1979; Prieditis, 1993; Culberson & Schaeffer, 1996). In particular, it has been shown that such heuristics can only be cost effective under certain conditions: the generalized theorem of Valtorta states that in the course of an A* search guided by a heuristic derived by searching blindly in some abstraction of the search space, every state that would be expanded by a blind search in the original search space must be expanded either in the abstract space or by the A* search in the original space (Holte, Perez, Zimmer, & MacDonald, 1996). This implies that if the abstraction is an embedding (the set of states in the abstract space is the same as in the original search space), such a heuristic can never be cost effective (Valtorta, 1984). The $m$-relaxation of the regression planning search space is an embedding, since every state in the normal regression space corresponds to exactly one state (containing the same set of subgoal atoms) in the $m$-regression space. In spite of this, there are reasons to believe that relaxed search can be cost effective: The algorithm used to search the $m$-regression space discovers (and stores in the heuristic table) the true $h^m$ value, or a lower bound on this value greater than that given by the current heuristic table, for every OR-node expanded during the course of the relaxed search. The AND/OR structure of the $m$-regression space, and the fact that the "on-line" heuristic makes use of all relevant information present in the heuristic table, implies that an improvement of the estimated cost of an OR-node may yield immediately an improved estimate of the cost of many AND-nodes (all states that are supersets of the improved state), without any additional search effort. Finally, because OR-nodes in the $m$-regression space are states of limited size, each node expansion in the $m$-regression space is likely to be computationally cheaper than the average in the normal regression space, since the number of successors generated when regressing a state generally increases with the number of goal atoms in the state.

## 3.2 IDAO*

To search the relaxed regression space, $\text{HSP}_a^*$ uses an algorithm called IDAO*. As the name suggests, it is an adaption of IDA* to searching AND/OR graphs, *i.e.*, it carries out a depth-first, iterative deepening search. IDAO* is admissible, in the sense that if guided by an admissible heuristic, it returns the optimal solution cost of the starting state. In fact, it finds the optimal cost of every OR-node that is solved in the course of the search. However, it does not keep enough information for the optimal solution itself to be extracted, so it can not be used to find solutions to AND/OR search problems. It works for the purpose of improving the heuristic, however, since for this only the optimal *cost* needs to be known.

```
(1)   IDAO*(s, b) {
(2)     solved = false;
(3)     current = h(s);
(4)     while (current < b and not solved) {
(5)        current = IDAO_DFS(s, current);
        }
(6)     return current;
      }

(7)   IDAO_DFS(s, b) {
(8)     if final(s) {
(9)        solved = true;
(10)       return 0;
        }
(11)    if (s stored in SolvedTable) {
(12)       solved = true;
(13)       return stored solution cost;
        }
(14)    if (|s| > m) {  // AND-node
(15)       for (each subset s' of s such that |s'| <= m) {
(16)         new cost of s' = IDAO*(s', b);  // call IDAO* with cost limit b
(17)         if (new cost of s' > b) {  // s' not solved
(18)            return new cost of s';
            }
          }
(19)       solved = (all subsets solved);
(20)       new cost of s = max over all s' [new cost of s'];
(21)       if (solved) {
(22)          store (s, new cost of s) in SolvedTable;
          }
(23)       return new cost of s;
        }
(24)    else {  // OR-node
(25)       for (each s' in succ(s)) {
(26)          if (delta(s,s') + h(s') <= b) {
(27)            new cost through s' = delta(s,s') + IDAO_DFS(s', b - delta(s,s'));
(28)            if (solved) {
(29)               new cost of s = new cost through s';
(30)               store (s, new cost of s) in SolvedTable;
(31)               return new cost of s;
             }
           }
(32)          else {
(33)            new cost through s' = delta(s,s') + h(s');
           }
         }
(34)       new cost of s = min over all s' [new cost through s'];
(35)       store (s, new cost of s) in HeuristicTable;
(36)       return new cost of s;
        }
      }
```

Figure 3: The IDAO* algorithm (with solved table).

The algorithm is sketched in Figure 3. The main difference from IDA* is in the DFS subroutine: when expanding an AND-node, it recursively invokes the main procedure IDAO*, rather than the DFS function. Thus, for each successor to an AND-node, the algorithm performs a series of searches with increasing cost bound, starting from the heuristic estimate of the successor node (which for some successors may be smaller than that of the AND-node itself) and finishing when a solution is found or the cost bound of the predecessor AND-node is exceeded. This ensures that the cost returned is always a lower bound on the optimal cost of the expanded node, and equal to the optimal cost if the node is solved. By storing updated costs of OR-nodes in the heuristic table, the search computes a part of the $h^m$ heuristic as a side effect and, as noted earlier, the values stored in the table become immediately available for use in subsequent heuristic evaluations. IDAO* stops searching the successors of an AND-node as soon as one is found to have a cost greater than the current bound, since this implies the cost of the AND-node is also greater than the bound. However, since the algorithm performs repeated depth-first searches with increasing bounds, remaining successors of the AND-node will eventually also be solved. When an $m$-solution has been found, all successors to every AND-node appearing in the solution tree have been searched, and their updated costs stored. This ensures that the resulting heuristic, *i.e.*, that defined by the heuristic table after the relaxed search is finished, is still consistent.

Because the successor nodes of AND-nodes are subsets, IDAO* frequently encounters the same state (set of goals) more than once during search. The algorithm can be speeded up, significantly, by storing solved nodes (both AND-nodes and OR-nodes) together with their optimal cost and short-cutting the search when it reaches a node that has already been solved[7]. In difference to the lower bounds stored in the heuristic table, which are valid also in the $m'$-regression search space for any $m' > m$ as well as in the original search space, the information in the solved table is valid only for the current $m$-regression search (since states of size $m'$, for $m' > m$ are relaxed in the $m$-regression space but not in the $m'$-regression space).

Note that a standard transposition table, which records updated cost estimates of unsolved nodes, is of no use in IDAO* since updated estimates of OR-nodes are stored in the heuristic table, while the heuristic estimate of an AND-node is always given by the maximum of its size $m$ successors.

### 3.2.1 AN EXAMPLE

For an illustration of the use of relaxed search to improve heuristic values, consider the following simple problem from the STRIPS version of the Satellite domain, introduced in the 2002 planning competition[8]. The problem concerns a satellite whose goal is to acquire images of different astronomical targets (represented by the predicate (`img ?d`)). To do so, its instrument must first be powered on ((`on`)) and calibrated ((`cal`)), and the satellite must turn so that it is pointing in the desired direction ((`point ?d`)). Instrument calibration requires the satellite to be pointing at a specific calibration target (in this example, direction

---

7. The solved table, like the transposition table, is implemented as a closed hashtable. In case of collisions, the previously stored node is simply overwritten. This means that some searches may be repeated, but does not affect correctness of the algorithm.

8. The domain used in this example is somewhat simplified. The full (temporal) domain is discussed in Section 4.3.

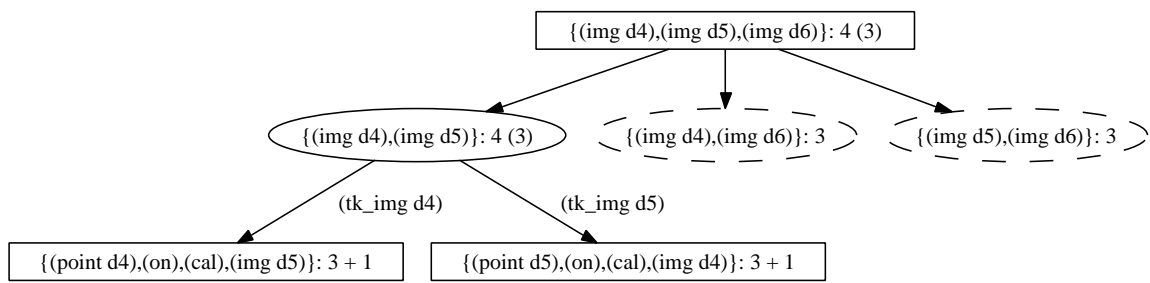

Figure 4: Part of the 2-Regression tree (expanded to a cost bound of 3) for the example Satellite problem. AND-nodes are depicted by rectangles, OR-nodes by ellipses. The cost of each node is written as "estimated + accumulated". For nodes whose estimated cost has been updated after expansion, the ($h^1$) estimate before expansion is given in parenthesis.

`d2`). Since this is the STRIPS version of the domain, all actions are assumed to have unit cost.

To keep size of the example manageable, let's assume a complete solution has been computed only for $h^1$ and that relaxed search is used to compute a partial $h^2$ solution. Figures 4 and 5 show (part of) the 2-relaxed space explored by the first and second iteration, respectively, of an IDAO* search starting from the problem goals.

In the first iteration (Figure 4) IDAO-DFS is called with a cost bound of 3, as this is the estimated cost of the starting state given by the precomputed $h^1$ heuristic. The root node is an AND-node, so when it is expanded IDAO* is called for each size 2 subset (lines $(15) - (18)$ in Figure 3). The first such subset to be generated is {(`img d4`),(`img d5`)}. This state also has an estimated cost of 3, so IDAO-DFS is called with this bound in the first iteration, but the two possible regressions of this state both lead to states with a higher cost estimate (an estimated cost of 3 plus an accumulated cost of 1). The new cost is propagated back to the parent state, where the improved cost estimate (4) of the atom set {(`img d4`),(`img d5`)} is stored in the heuristic table and returned (lines $(35) - (36)$ in Figure 3). Since this puts the estimated cost of the state now above the bound of the IDAO* call (line (4) in Figure 3) no more iterations are done. The new cost is returned to the IDAO-DFS procedure expanding the root node, which also returns since the root node is an AND-node and it now has an unsolved successor (lines $(17) - (18)$ in Figure 3). This finishes the first iteration.

In the second iteration (Figure 5) IDAO-DFS is called with a bound of 4. It proceeds like the first, but now the estimated cost of the AND-node {(`point d4`),(`on`),(`cal`),(`img d5`)} is within the bound, so this node is expanded. The first size 2 subset for which IDAO* is called is {(`point d4`),(`on`)}, with an initial estimated cost of 1. The first iteration fails to find a solution for this state, but since the new cost of 2 is still within the bound imposed by the parent AND-node, a second iteration is done which finds a solution. The new cost of the atom set {(`point d4`),(`on`)} is stored in the heuristic table and in addition, the solved states (along with their optimal solution cost) are all stored in the solved table (lines

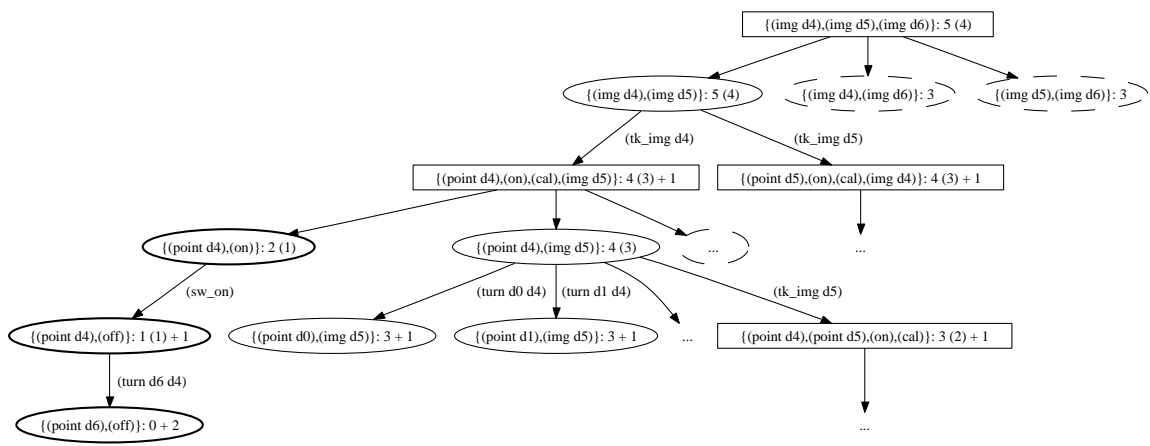

Figure 5: Part of the 2-Regression tree (expanded to a cost bound of 4) for the example Satellite problem. AND-nodes are depicted by rectangles, OR-nodes by ellipses. The cost of each node is written as "estimated + accumulated". Note that the accumulated cost is only along the path from the nearest ancestor AND-node. For nodes whose estimated cost has been updated after expansion, the estimate before expansion is given in parenthesis: this estimate includes updates made in the previous iteration (shown in Figure 4).

(28) − (31) in Figure 3). Since the first successor of the AND-node was solved expansion continues with the next subset, {(point d4),(img d5)}. This state has several possible regressions, some of which lead to OR-nodes but some to AND-nodes. All, however, return a minimum (estimated + accumulated) cost of 4, so an improved cost (for the atom set {(point d4),(img d5)}) is stored in the heuristic table and the parent AND-node remains unsolved. A similar process happens when its sibling node, {(point d5),(on),(cal),(img d4)}, is expanded, and the cost of the atom set {(img d4),(img d5)} is updated once more, to 5.

The process continues through a few more iterations, until all the size 2 subsets of the top-level goal set have been solved, the most costly at a cost of 7. At this point, updated estimates of 65 size 2 atom sets have been stored in the heuristic table, slightly less than half the number that would have been stored if a complete $h^2$ solution had been computed.

## 3.3 Relaxed Search: Temporal Case

As was the case with the $h^m$ heuristic itself, adapting relaxed search to the temporal case is simple in principle, but somewhat complicated in practice. First, the relaxation introduced by equations (4) − (5) approximates a temporal regression state $s = (E, F)$ by a *set* of states without actions, *i.e.*, of the form $(E', \emptyset)$. To keep matters simple, only equation (5) is used

```
TP4(problem) {                      HSP*a(problem) {
   solve h^2 by GBF, store in HeuristicTable;    solve h^2 by GBF, store in HeuristicTable;
   opt = IDA*(problem.goals);       m = 3;
}                                   while (not <stopping condition>) {
                                       IDAO*(problem.goals, infinity);
                                       if (m-relaxed problem not solved) {
                                          fail; // original problem unsolvable
                                       }
                                       m = m + 1;
                                    }
                                    opt = IDA*(problem.goals);
                                 }
```

Figure 6: The TP4 and HSP$_a^*$ planning procedures.

in the relaxed search: that is, the size of a state is defined as

$$|(E, F)| = \left| E \cup \bigcup_{(a,\delta) \in F} pre(a) \right|. \tag{6}$$

Second, even so a state of size less than $m$ may still have a non-empty $F$ component, and such a state can not be stored in the heuristic table (which maps only atom sets to associated costs). Neither can the optimal cost or lower bound found for such a state be stored as the cost of the corresponding atom set (right hand side of equation (6)), since the optimal cost of achieving this atom set may be lower. However, a plan that achieves $E \cup \bigcup_{(a,\delta) \in F} pre(a)$ at $t$ also achieves the state $(E, F)$ at most $\max_{(a,\delta) \in F} \delta$ time units later, through inertia, *i.e.*,

$$h^*(E, F) \leqslant h^* \left( E \cup \bigcup_{(a,\delta) \in F} pre(a), \emptyset \right) - \left( \max_{(a,\delta) \in F} \delta \right) \tag{7}$$

Thus, to maintain the admissibility of the heuristic function defined by the contents of the heuristic table, the largest $\delta$ among all actions in $F$ is subtracted from the cost before it is stored.

Unfortunately, both of these simplifications weaken the heuristic values found by relaxed search. What is worse, since a cost under-approximation is applied when storing states containing concurrent actions, but not during the search, the heuristic defined by the table after relaxed search can be inconsistent. Also, right-shift cuts can not be used in the relaxed search. As mentioned earlier, in the search space pruned by right-shift cuts, the possible successors to a state, and therefore the cost returned when the state is expanded (regressed) but not solved, may be different depending on the path through which it was reached. Again, this can not be stored in the heuristic table, and it can not be ignored since this could violate the admissibility of the heuristic.

### 3.4 HSP$_a^*$

The HSP$_a^*$ planning procedure (shown in Figure 6) consists of three main steps: the first is to precompute the temporal $h^2$ heuristic, the second to perform a series of $m$-relaxed searches, for $m = 3, \ldots$, in order to improve the heuristic, and the final is an IDA* search in the temporal regression space, guided by the computed heuristic. Note that the first and last of the three steps are identical to those of TP4: the only difference is the intermediate step, the series of relaxed searches. The purpose of these searches is to discover, and store in the heuristic table, improved cost estimates of states (*i.e.*, atom sets) of size $m$. As indicated in Figure 6, relaxed searches are carried out for $m = 3, \ldots$, until some stopping condition is satisfied. There are several reasonable stopping conditions that can be used:

(*a*) stop when the last $m$-regression search does not encounter any AND-node (in which case the relaxed solution is in fact a solution to the original problem);

(*b*) stop at a fixed *a priori* given $m$;

(*c*) stop when the cost of the $m$-solution found is the same as that of the $(m-1)$-solution (or heuristic estimate);

(*d*) stop after a certain amount of time, number of expanded nodes, or similar.

HSP$_a^*$ implements the first three. Each results in a different configuration of the planner, and usually also in a difference in performance. In the competition, a fixed limit at $m = 3$ was used. Except where it is explicitly stated otherwise, this is the configuration used in the experiments presented in the next section as well.

## 4. Results in the Competition Domains

This section presents a comparison of the relative performance of TP4 and HSP$_a^*$ on the domains and problem sets that were used in the 2004 planning competition, and an analysis of the results. The results presented here are from rerunning both planners on the competition problem sets, not the actual results from the competition. This is for two reasons: First, as already mentioned, errors in the HSP$_a^*$ implementation made its performance in the competition somewhat worse than what it is actually capable of. Second, the repeated runs were made with a more generous time limit than that imposed during the competition to obtain more data and enable a better comparison[9]. Also, some experiments were run with alternative configurations of the planners. Detailed descriptions of the competition domains are given by Hoffmann, Edelkamp *et al.* (2004, ?).

### 4.1 The pipesworld Domain

The pipesworld domain models transportation of "batches" of petroleum products through a pipeline network. The main difference from other transportation domains is that the

---

9. The experiments were made with CPU time limits between 4 and 8 hours for each problem, though on slightly a slower machine than that used in the competition: a Sun Enterprise 4000 which has 12 processors at 700 MHz and 1024 MB memory in total. The multiple processors offer no advantage to the planners, since these are of course single-threaded, but are used to run several instances in parallel, shortening the overall "makespan" of the experiment.

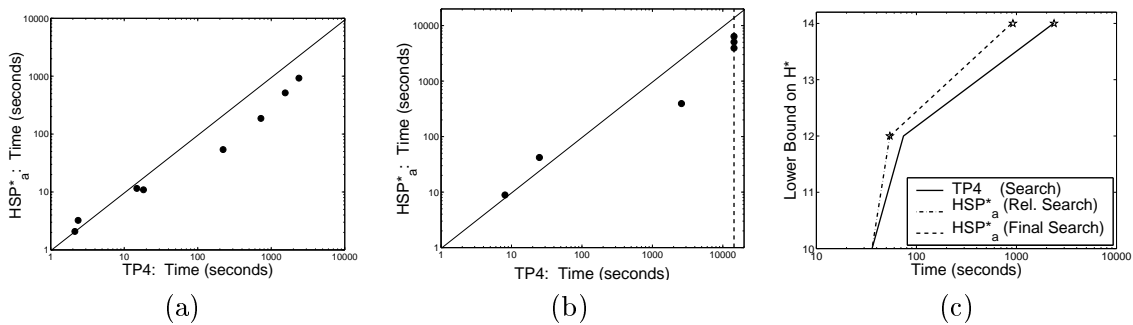

(a)                    (b)                    (c)

Figure 7: Solution times for TP4 and $\text{HSP}_a^*$ on problems solved in the `pipesworld` domain
(a) without tankage restriction, and (b) with tankage restrictions. The vertical
line to the right in figure (b) indicates the time-out limit (thus, the three points
on the line correspond to problem instances solved only by $\text{HSP}_a^*$). (c) Evolution of
the lower bound on solution cost during relaxed and normal (non-relaxed) search
in problem `p08` of the `pipesworld` domain (version without tankage restriction).
Stars indicate where solutions are found. Note that all time scales are logarithmic.

pipelines must be filled at all times, so when one batch enters a pipe (is "pushed") another
batch must leave the pipe at the other end (be "popped"). The domain comes in two versions, one with restrictions on "tankage" (space for intermediary storage) and one without
such restrictions.

Although neither TP4 nor $\text{HSP}_a^*$ achieve very good results in this domain, it is an example
of a domain where $\text{HSP}_a^*$ performs better than TP4. Figures 7(a) and 7(b) compare the
runtimes of the two planners on the set of problems solved by at least one.

Figure 7(c) compares the behaviour of the two planners on one example problem, `p08`
from the domain version without tankage restriction, in more detail. This provides an
illustrative example of relaxed search when it works as intended. Since both planners use
iterative deepening searches, the best known lower bound on the cost of the problem solution
will be increasing, starting from the initial $h^2$ estimate, through a series of (relaxed and
non-relaxed) searches with increasing bound, until a solution is found: the graph plots this
evolution of the solution cost lower bound against time. As can be seen, 3-regression search
reaches a solution (with cost 12) faster than the normal regression search discovers that
there is no solution within the same cost bound. The final (non-relaxed) regression search
in $\text{HSP}_a^*$ is also faster than that of TP4 (as indicated by the slope of the curve), due to the
heuristic improvements stored during the relaxed search.

## 4.2 The `promela` and `psr` Domains

Certain kinds of model checking problems, such as the detection of deadlocks and assertion
violations, are essentially questions of reachability (or unreachability) in state-transition
graphs. The `promela` domain is the result of translating such model checking problems, for
system models expressed in the *Promela* specification language, into PDDL. The problems used in the competition are instances of two different deadlock detection problems (the "dining philosophers" and "optical telegraph" problems) of increasing size.

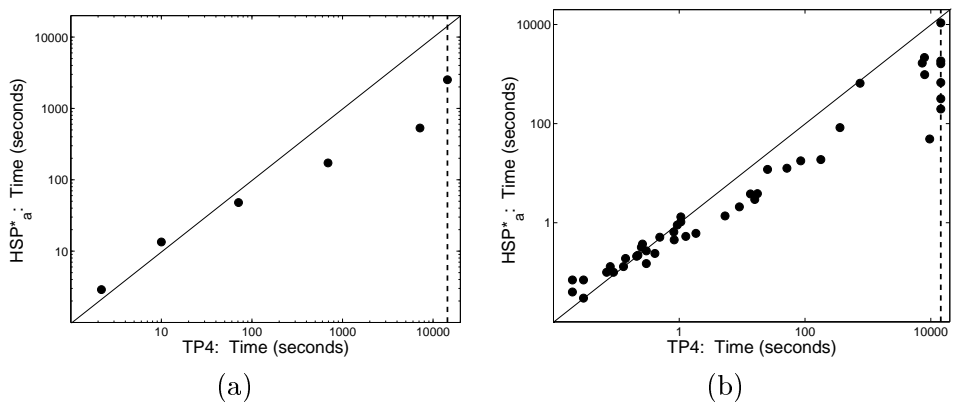

(a)          (b)

Figure 8: Solution times for TP4 and $\text{HSP}^*_a$ on problems solved in (a) the `promela` domain (`philosophers` subset) and (b) the `psr` domain (`small` instance subset). The vertical line to the right indicates the time-out limit (thus, points on the line correspond to problem instances solved only by $\text{HSP}^*_a$).

The `psr` domain models the problem of reconfiguring a faulty power network to resupply consumers affected by the fault. Uncertainty concerning the initial state of the problem (the number and location of faults), unreliable actions and partial, sometimes even false, observations are important features of the application, but these aspects were simplified away from the domain used in the competition. The domain did however make significant use of ADL constructs and the new derived predicates feature of PDDL2.2. The ADL constructs and derived predicates can be compiled away, but only at an exponential increase in problem size. Therefore only the smallest instances were available in plain STRIPS formulation, and because of this they were the only instances that TP4 and $\text{HSP}^*_a$ could attempt to solve.

The `promela` and `psr` domains are non-temporal, in the sense that action durations are not considered, but neither are they strictly sequential, *i.e.*, actions can take place concurrently. Because of this, TP4 and $\text{HSP}^*_a$ were run in parallel, rather than temporal, planning mode on problems in these domains[10]. The results of the two planners, shown in Figure 8, are similar to those exhibited in the `pipesworld` domain: $\text{HSP}^*_a$ is better than TP4 overall, solving more problems in both domains and solving the harder instances faster, while TP4 is faster at solving easy instances.

### 4.3 The `satellite` Domain

The `satellite` domain models satellites tasked with making astronomical observations. A simplified STRIPS version of the domain was described in Section 3.2.1. In the general do-main, there can be more than one satellite, each equipped with more than one instrument,
and different instruments have different imaging capabilities (called "modes"), which may
overlap (between instruments and between satellites). Each goal is to have an image taken
of a specific target, in a specific mode. As in the STRIPS version, taking an image requires
the relevant instrument to be powered on and calibrated, and to calibrate an instrument
the satellite must be pointed towards a calibration target. Turning times between differ-
ent directions vary. As an additional complication, at most one instrument onboard each
satellite can be powered on at any time. Thus, to minimize overall execution time requires
a careful selection of which satellite (and instrument) to use for each observation, and the
order in which each satellite carries out the observations it has been assigned.

---

10. Parallel planning is the special case of temporal planning that results when all actions have unit durations. Certain optimizations for this case are implemented (identically) in both planners.

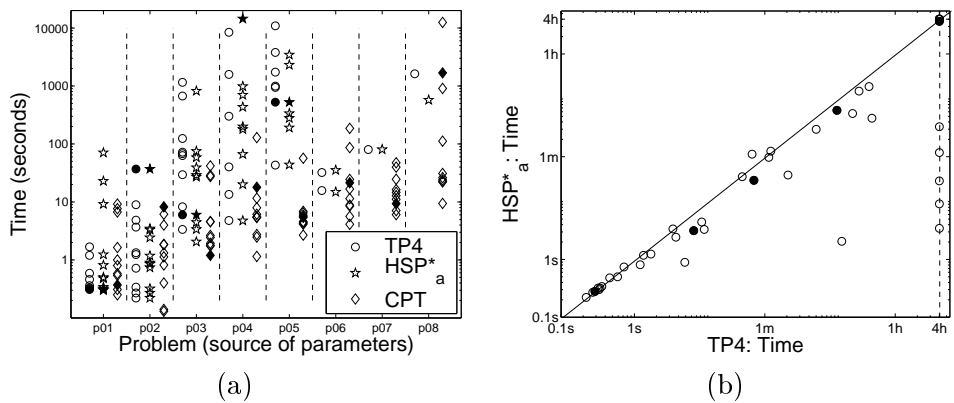

(a)                                        (b)

Figure 9: Solution times for TP4 and $\text{HSP}^*_a$ on problems solved in the `satellite` domain.
Filled (black) points represent instances belonging to the competition problem
set, while remaining points are from the set of additional problems generated.
Each "wide" column in figure (a) represents one problem from the competition
set and shows the solution times for the set of instances generated with the same
parameters (grouped into subcolumns by planner). Only solved instances are
shown, so not all columns have the same number of points. Figure (b) compares
TP4 and $\text{HSP}^*_a$ directly. The vertical line to the right indicates the time-out limit
(thus, points on the line are instances solved by $\text{HSP}^*_a$ but not by TP4).

This domain is hard for both TP4 and $\text{HSP}^*_a$, for several reasons: First, as already
mentioned, the core of the domain is a combination an assignment problem and a TSP-like
problem, both of which are hard optimization problems. Also, the $h^2$ heuristic tends to be
particularly weak on TSP and related problems (the same weakness has also been noted by
Smith (2004) for the planning graph heuristic, which is essentially the same as $h^2$). Second,
action durations in this domain differ by large amounts and are at the same time specified
with a very high resolution. For example, in problem `p02` one action has a duration of
1.204 and another a duration of 82.99. When using IDA* with temporal regression, the
cost bound tends to increase by the gcd (greatest common divisor) of action durations in each iteration, except for the first few iterations[11]. In the `satellite` domain, the gcd of action durations is typically very small (on the order of $\frac{1}{100}$). Combined with the weakness of the $h^2$ heuristic, which means the difference between the initial heuristic estimate of the solution cost (makespan) of a problem and the actual optimal cost is often large, this results in an almost astronomical number of IDA* iterations being required before a solution is found. To avoid this (somewhat artificial) problem, action durations were rounded up to the nearest integer in the experiments done in this domain. This increases the makespan of the plans found, but not very much – on average by 2.9%, and at most by 5.9% (comparison made on the problems that could be solved with original durations)[12].

Due to the weakness of the $h^2$ heuristic in this domain, the effort invested by HSP$_a^*$ in computing a more accurate heuristic can be expected to pay off, resulting in a better overall runtime for HSP$_a^*$ compared to TP4. This is indeed the case: although HSP$_a^*$ solves only the five smallest problems in the set (shown as black points in Figure 9(b)), TP4 solves only four of those, and is slightly slower on most of them. These results, however, are not quite representative.

The `satellite` domain has a large number of problem parameters: the number of goals and the number of satellites, instruments and the instrument capabilities, *etc.*, which determine the number of ways to achieve each goal. Problem instances used in the competition were generated randomly, with varying parameter settings[13]. The competition problem set, which has to offer challenging problems to a wide variety of planners (both optimal and suboptimal) while for practical reasons not being too large, scales up the different parameters quite steeply, and – more importantly – contains only one problem instance for each set of parameters used. However, the hardness of a problem instance may depend as much (if not more) on the random elements of the problem generation (which include, *e.g.*, the turning times between targets and the actual allocation of capabilities and calibration targets to instruments) as on the settings of the controllable parameters. To investigate the importance of the random problem elements for problem hardness, and to obtain a broader basis for the comparison between TP4 and HSP$_a^*$, ten additional problems were generated (using the available problem generator) for each of the parameter settings corresponding to the eight smallest problems in the competition set. The distribution of solution times for TP4, HSP$_a^*$ and CPT (the only optimal temporal planner besides TP4 and HSP$_a^*$ to partici-pate in the competition) on each of the resulting problem sets is shown in Figure 9(a). The instances that were part of the competition problem set are shown by filled (black) points. Clearly, the variation in problem hardness is considerable and of the problems in the competition set some are very easy and some are very hard, relative to the set of problems generated with the same parameters.

---

11. TP4 and HSP$_a^*$ treat action durations as rationals: by the gcd of two rationals $a$ and $b$ is meant the greatest rational $c$ such that $a = mc$ and $b = nc$ for integers $m$ and $n$. Note that the planners do *not* compute the gcd of action durations and use this to increment the cost bound. The bound is in each iteration increased to the cost of the least costly node that was not expanded due to having a cost above the bound in the previous iteration (as per standard IDA* search). That this frequently happens to be (on the order of) the gcd of action durations is an (undesirable) effect of the branching rule used to generate the search space.

12. Optimality can be restored by a two-stage optimization scheme, in which the makespan of the non-optimal solution is taken as the initial upper bound in a branch-and-bound search, using the original action durations (see Haslum, 2004b, for more detail). This was used in the competition for the `satellite` domain, where the two search stages combined take less time than a plain IDA* search using original durations. The two-stage scheme is applicable to any domain, but its effectiveness in general is an open question.

13. The problem generator can be found at http://planning.cis.strath.ac.uk/competition/. The controllable parameters are the number of satellites, the maximum number of instruments per satellite, the number of different observation modes, the total number of targets, and the number of observation goals.

Figure 9(b) compares TP4 and $\text{HSP}_a^*$ on the extended problem set. $\text{HSP}_a^*$ solves 59% of this set, while TP4 solves 51% (a subset of those solved by $\text{HSP}_a^*$). However, as can be seen in the figure, the relative performance of the two planners is also highly varied, much more so than the results on the competition problem set suggests.

## 4.4 The `airport` Domain

The `airport` domain models the movements of aircraft on the ground at an airport. The goal is to guide arriving aircraft to parking positions and departing aircraft to a suitable runway for takeoff, along the airport network of taxiways. The main complication is to keep the aircraft safely separated: at most one aircraft can occupy a runway or taxiway segment at any time, and depending on the size of the aircraft and the layout of the airport nearby segments may be blocked as well.

TP4 solves only 13 out of the 50 problem instances in this domain. For the instances solved by TP4, the number of nodes expanded in search is very small relative to the solution depth (though for the larger instances, node expansion is very slow, resulting in a poor runtime overall). This implies that for these problem instances the $h^2$ heuristic is very accurate, and thus they are in a sense "easy"; for such instances, $\text{HSP}_a^*$ can not be expected to be better, since the search effort it invests into computing a more accurate heuristic is largely wasted, but it also indicates that a more accurate heuristic is needed to solve "hard" problem instances.

However, $\text{HSP}_a^*$ solves only 7 problems, a subset of those solved by TP4, and takes far more time for each. Figure 10(a) shows the time $\text{HSP}_a^*$ spends in 3-regression search and in the final (non-relaxed) search for each of the `airport` instances it solves. For reference, the search time for TP4 is also included. Clearly, the relaxed search consumes a lot of time in this domain, and offers very little in the way of heuristic improvement in return. That the heuristic improvement is small (close to non-existent) is easily explained, since, as already observed, the $h^2$ heuristic is already very accurate on these particular problem instances. The question, then, is why the relaxed search is so time consuming.

The apparent reason is that in this domain, search in the 3-regression space is more expensive than search in the normal regression space. This is contrary to the assumption stated in Section 3.1, that the cost of expanding a state should be smaller in the relaxed regression space, due to a smaller branching factor. Table 10(b) displays some characteristics of the normal and 3-regression spaces for `airport` instance `p08` (the smallest instance not solved by TP4). Data is collected during the first (failed) iteration of IDA*/IDAO*. States in the normal regression space contain, on average, a large number of subgoals, while in the 3-regression space, states corresponding to OR-nodes are by definition limited in size. Consequently, the branching factor of OR-nodes in 3-regression is smaller (since the choice of establisher for each subgoal is a potential branch point), but not by much: the many subgoals in the normal regression interact, resulting in relatively few consistent choices.

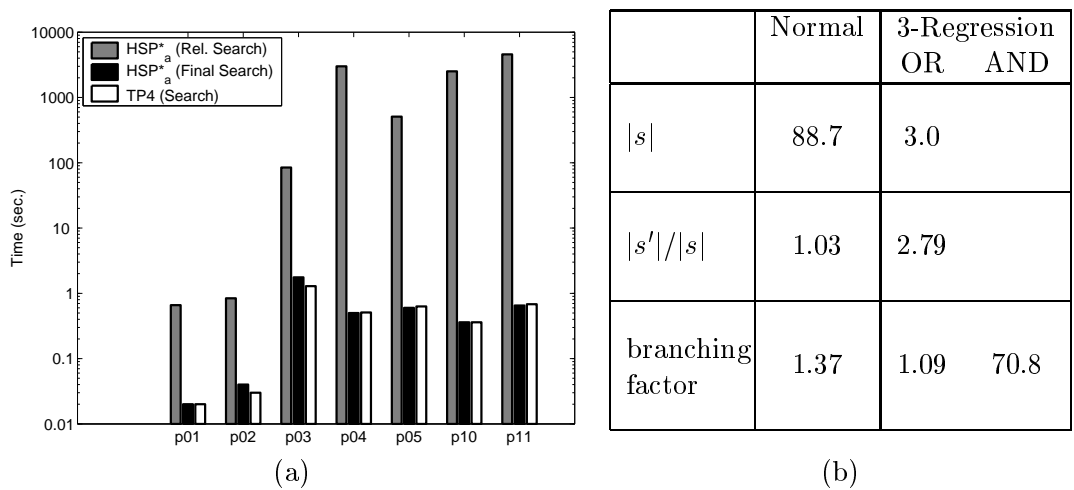

| | Normal | 3-Regression | |
|---|---|---|---|
| | | OR | AND |
| $|s|$ | 88.7 | 3.0 | |
| $|s'|/|s|$ | 1.03 | 2.79 | |
| branching factor | 1.37 | 1.09 | 70.8 |

(a)                                             (b)

Figure 10: (a) Time spent in 3-regression search and in final (non-relaxed) search on airport instances solved by HSP$_a^*$. The search time for TP4 is also shown for comparison. Note the logarithmic time scale: search times for HSP$_a^*$ and TP4 are nearly identical, while the 3-regression search consumes several orders of magnitude more time. (b) Characteristics of the normal and 3-regression spaces for airport instance p08: $|s|$ is the average state size; $|s'|/|s'|$ the average ratio of successor state size to the size of the predecessor state. Data is collected during the first (failed) iteration of IDA*/IDAO*.

Also, the right-shift cut rule, which eliminates some redundant branches, is used in the normal regression space, but not when expanding OR-nodes in 3-regression. However, regression tends to make states "grow", *i.e.*, successor states generally contain more subgoals than their predecessors, and while this effect is quite moderate in normal regression, where successors have, on average, 3% more subgoals, it is much more pronounced for the smaller states corresponding to OR-nodes in the 3-regression space, whose successors are on average 2.79 *times* larger. As a result, successors to OR-nodes are all AND-nodes, with an average of about 8.3 subgoals and 70.8 successors (subsets of size 3).

To summarize, each expanded OR-node in 3-regression results (via an intermediate "layer" of AND-nodes) in an average of 77.2 new OR-nodes. Even though most of them (74.2%) are found in the IDAO* solved table, and therefore don't have to be searched, those that remain yield an effective "OR-to-OR" branching factor of 19.9 (25.8% of 77.2), to be compared with the branching factor of 1.37 for normal regression. Again, the problem is not the high branching factor in itself: it is that the branching factor in the relaxed search space is far *higher* than it is for normal regression, and that search in the 3-regression space is consequently more expensive than search in the normal regression space, rather than less.

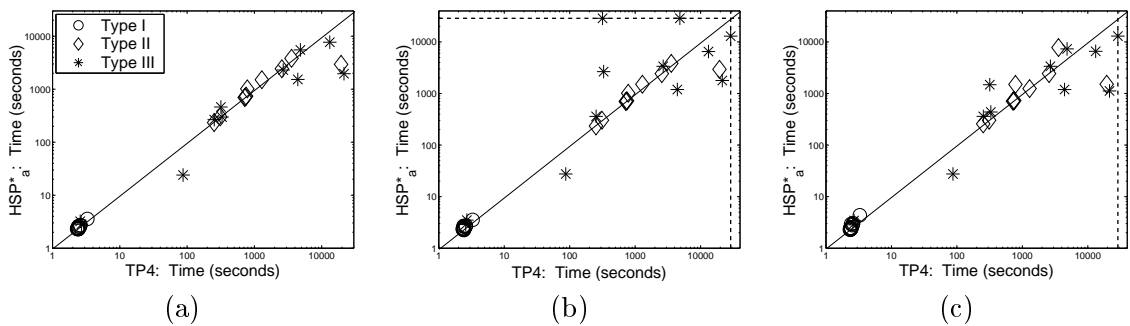

(a) (b) (c)

Figure 11: Solution times for TP4 and three different configurations of HSP$_a^*$ on problems solved in the `umts` domain: (a) HSP$_a^*$ with $m$-regression limited to $m = 3$ only; (b) "unlimited" HSP$_a^*$ (performs $m$-regressions for increasing $m$ until either a non-relaxed solution is found, or the estimated cost of the top level goals does not increase); (c) "3–4" HSP$_a^*$ (always performs 3- and 4-regression). The lines to the right and top in figures (b) and (c) indicate the time-out limit. "Type I – III" refers to the classification of the problem instances described in Section 4.5 (page 257).

## 4.5 The `umts` Domain

The `umts` domain models the UMTS call set-up procedure for data applications in mobile telephones. The domain is actually a scheduling problem, similar to flowshop. The call set-up procedure consists in eight discrete steps for each application, ordered by precedence constraints. The duration of a step depends on the type of step as well as the application. When several applications are being set up, steps pertaining to different applications can be executed in parallel, subject to resource availability: there are 15 numeric resources, and each step uses a certain amount, which depends on the application, of some subset of resources during execution (only 3 of the 15 resources are actually oversubscribed). Resources are "reusable", *i.e.*, the amount used becomes available again once the step has finished.

At first glance, this appears to be a perfect domain for HSP$_a^*$: The presence of reusable resource limitations makes it more likely that there are higher-order mutual exclusions between actions (*i.e.*, there may be enough of a resource to carry out two actions using the resource concurrently, but not three, or three but not four, *etc*). This suggests the $h^m$ heuristics are more likely to improve with increasing $m$, since $h^m$ considers at most $m$ subgoals and therefore at most $m$ concurrent actions. At the same time, due to the simple structure of precedence constraints, the "growing" states and the resulting branching factor blow-up in relaxed search that occurred in the `airport` domain, are unlikely.

Results, however, disagree: TP4 and HSP$_a^*$ solve the same set of problem instances (39 out of 50), in times as shown in Figure 11(a). TP4 is faster than HSP$_a^*$ in a majority of cases, though the difference is relatively small, while in the cases where HSP$_a^*$ is the fastest of the two, the difference is greater. To see why, the problem instances can be divided into the following three types:

**Type I**: Instances in which all contested resources are available in sufficient quantity. This means there is no resource conflict at all, and thus that $h^* = h^1$. There are 18 instances of this type among those solved, and they are indicated by "○" in Figure 11.

**Type II**: Instances in which $h^3 = h^2$ but $h^* > h^2$, *i.e.*, there are resource conflicts, but these involve more than three concurrent actions and are therefore not detected by $h^3$. There are 10 instances of this type among those solved, and they are indicated by "◇" in Figure 11.

**Type III**: Instances in which $h^3 > h^2$. There are 11 instances of this type among those solved, and they are indicated by "∗" in Figure 11.

On type I instances, HSP$_a^*$ clearly pays an overhead for computing an unnecessarily strong heuristic, though it is relatively small. Note that these account for a third of the instances in the problem set (18 out of 50), and nearly half of the solved instances. On instances of types II and III, HSP$_a^*$ expands fewer nodes in the final (non-relaxed) search than TP4 does during its search, as much as 34% fewer on average. This shows that the heuristic improvement resulting from 3-regression is at least of some value, though in roughly $\frac{1}{3}$rd of the instances not enough to compensate for the cost of performing the relaxed search.

Recall that the HSP$_a^*$ planner in the competition, and in the experiments presented here so far, was restricted to performing only 3-regression search. A possible explanation for the relatively weak results the planner produces on type II and III instances is that this restriction prevents relaxed search from being fully exploited. However, this theory does not hold. Figures 11(b) and 11(c) show results for two alternative configurations of HSP$_a^*$: in 11(b) an "unlimited" configuration, which carries out $m$-regression searches for $m = 3, \ldots$ until either a non-relaxed solution is found, or the optimal $m$-solution cost is found to be the same as the $(m-1)$-solution cost, and in 11(c) a "3–4" configuration, which always performs 3- and 4-regression searches. As can be seen in the figures, both alternative configurations incur a larger overhead for the relaxed searches (unlimited HSP$_a^*$ even times out on two instances while doing 5-regression search). Also, the gain from the additional heuristic information is quite small: comparing again the number of nodes expanded in the final (non-relaxed) regression search to the number of nodes expanded by TP4, the unlimited and 3–4 configurations expand on average 41.4% and 41% fewer nodes, respectively (to be compared to the 34% average saving in expanded nodes obtained by HSP$_a^*$ restricted to 3-regression search only).

Another possible explanation is that the transposition table, which also stores updates of estimated cost, though for whole states rather than subsets of goals, to some extent compensates for the weaker heuristic used by TP4. Again, however, the explanation turns out not to hold: with the transposition table disabled in both planners, the savings in number of nodes expanded in the final search by HSP$_a^*$ compared to the number of nodes expanded by TP4 on type II and III instances is actually less than whend transposition tables are used, averaging only 18% (though HSP$_a^*$ in this experiment solves two problems that TP4 fails to solve), and the difference also becomes much more varied.

Recall that a number of simplifications were introduced in the formulation of temporal $m$-regression, in order to enable complete solutions to the relaxed cost equation to be computed and stored in the heuristic table. Thus, a remaining possible explanation for the small value of the improvement of the heuristic in the final search is these simplifications, since they lower the estimates stored in the heuristic table.

Concerning the time overhead for relaxed search, in particular for the higher $m$-regression searches, the explanation appears again to be a higher branching factor in the relaxed search space, though the situation is somewhat different than in the `airport` domain. The ratio between the size of successor states and their predecessors is low, averaging 0.97 in normal regression and 0.91 for OR-nodes in 3-regression (and stays roughly the same also in the 4- and 5-regression spaces) so AND-nodes are relatively scarce. But the average branching factor for OR-nodes in 3-regression is 2.4 (increasing to 4.98 and 8.27 in 4- and 5-regression, respectively) compared to an average of 1.85 in normal regression. The reason in the `umts` domain is the right-shift cuts: recall that these eliminate redundant branches from the search space, thus reducing the branching factor, but can not be used when regressing OR-nodes in relaxed search since this might cause the computed heuristic to become inadmissible. The branching factor for normal regression search without right-shift cuts is 2.69. The difference may seem small, but it has a great effect: TP4 without right-shift cuts fails to solve all but two type II and III instances (in many cases not even finishing the first DFS iteration).

## 4.6 Analysis and Conclusions

In several of the competition domains, HSP$_a^*$ does achieve better results than TP4, indicating that relaxed search can be an efficient method of computing a more accurate heuristic while staying in the $h^m$ framework. In the domains where it fails – the `airport` and `umts` domains – it does so because relaxed search yields a relatively small improvement over the $h^2$ heuristic, at a disproportionately large computational cost.

In the `airport` domain, the two problems are tightly connected: the heuristic improvement is negligible simply because relaxed search is so expensive that the only problem instances on which it finishes within the time limit are those very simple instance for which the $h^2$ heuristic is already close to perfect. In the `umts` domain, the reason for the poor heuristic improvement is still somewhat of a mystery: a number of hypotheses were tested, and refuted. A remaining plausible explanation is that the simplifications introduced in formulation of temporal $m$-regression are particularly damaging in this domain.

In both domains, however, the explanation for the relatively large overhead for relaxed search appears to be that it suffers from a higher branching factor than the normal regression search, which causes expansion of OR-nodes in the relaxed search to be computationally more expensive than node expansion in the normal search (even if many of the generated successors are not searched). Figure 12 summarizes some search space characteristics for all the competition domains. In the domains where relaxed search is expensive, this is because states tend to grow when regressed, $i.e.$, $|s'|/|s|$ is large and as a result there are many AND-nodes with many successors (`airport`), or because OR-nodes in the relaxed regression space have a higher branching factor than in the normal regression space, and are therefore computationally more expensive to expand (`umts`). In the domains where relaxed search is successful, on the other hand, $|s'|/|s|$ is typically close to 1, $i.e.$, small states stay small when regressed, and regression of OR-nodes in the relaxed space is computationally cheaper than node expansion in the normal regression space, as indicated by a lower (or roughly equal) branching factor. Of course, these (averaged) numbers are not perfect predictors of performance: in the `promela` domain, for example, the $|s'|/|s|$ ratio is quite large but $\text{HSP}_a^*$ outperforms TP4 anyway (as shown in Figure 8(a)).

| | Normal Regression | 3-Regression | |
|---|---|---|---|
| | | OR | AND |
| **airport** | | | |
| $\lvert s \rvert$ | 88.7 | 3.0 | |
| $\lvert s' \rvert / \lvert s \rvert$ | 1.03 | 2.79 | |
| branching factor | 1.37 | 1.09 | 70.8 |
| **pipesworld** | | | |
| $\lvert s \rvert$ | 6.76 | 2.99 | |
| $\lvert s' \rvert / \lvert s \rvert$ | 1.21 | 1.35 | |
| branching factor | 15.1 | 5.13 | 5.2 |
| **promela (philosophers)** | | | |
| $\lvert s \rvert$ | 14.9 | 2.99 | |
| $\lvert s' \rvert / \lvert s \rvert$ | 1.17 | 2.17 | |
| branching factor | 21.2 | 3.30 | 30.6 |
| **psr** | | | |
| $\lvert s \rvert$ | 9.05 | 2.99 | |
| $\lvert s' \rvert / \lvert s \rvert$ | 1.08 | 1.42 | |
| branching factor | 24.5 | 15.1 | 7.75 |
| **satellite** | | | |
| $\lvert s \rvert$ | 6.88 | 2.99 | |
| $\lvert s' \rvert / \lvert s \rvert$ | 1.04 | 1.06 | |
| branching factor | 6.98 | 5.01 | 7.33 |
| **umts** | | | |
| $\lvert s \rvert$ | 8.10 | 2.52 | |
| $\lvert s' \rvert / \lvert s \rvert$ | 0.97 | 0.91 | |
| branching factor | 1.85 | 2.40 | 6.36 |

Figure 12: Some characteristics of the normal regression and 3-regression search spaces in the domains considered: the average state size ($\lvert s \rvert$), the average ratio of successor state size to the size of the predecessor state ($\lvert s' \rvert / \lvert s \rvert$) and the branching factor. For the `pipesworld`, `promela`, `psr` and `satellite` domains, the numbers shown are the averages over solved problem instances. For the `umts` domain, the average is over solved type II and III instances only (see Section 4.5). For the `airport` domain, data is from a single (failed) iteration on a single problem instance (`p08`).

It is instructive to look more closely at the states in the `airport` domain, and why they grow when regressed. For example, the "state" of each aircraft is (in each world state) described by three components: one tells where the aircraft is positioned in the network of airport runways and taxiways, one which direction it is facing, and one whether it is parked, being pushed or moving under its own power[14]. Almost every operator that changes one of these has an effect or precondition on the other two as well. For example, any instance of the `move` operator, which changes the position of an aircraft, requires the aircraft to be moving and facing a particular direction, and may also change the facing. Thus, regressing a state containing only a goal atom belonging to one of the components in most cases results in a state containing goal atoms belonging to all three. A conclusion one may draw is that splitting large states (AND-nodes) into smaller states (OR-nodes) based only on the number on atoms is not always the right choice. An alternative would be to divide atoms in the planning problem into groups of "related" atoms and take the number of groups represented in a state to be its size.

Another observation that can be made is that the bulk of time spent in relaxed search is spent in the final search iteration, when a solution exists within the current cost bound. This iteration is not only the most expensive, but also the least useful, since relatively few heuristic improvements are discovered in it. This also relates to the branching factor, specifically the fact that AND-nodes have many more successors than OR-nodes: for an AND-node to be solved all its successors must be solved, so in the final iteration, all successors of every AND-node are searched. However, the purpose of relaxed search is not to find a solution in the $m$-regression space, but to find size $m$ states (OR-nodes) whose cost is underestimated by the heuristic, and more accurate cost estimates for these. Therefore it may not actually be necessary to search all the successors to every AND-node. An alternative would be a $(m, k)$-regression search in which only the $k$ most "promising" successors to every AND-node are considered, yielding another dimension for iteratively refining the heuristic (this is discussed further in the next section).

Finally, it should be pointed out that even though $\text{HSP}_a^*$ is, on average, better than TP4, results of both planners on the competition domains still appear rather poor. The only other optimal temporal planner to participate in the competition was CPT (Vidal & Geffner, 2004), which, in most domains, achieved much better results than TP4 (the results on the extended `satellite` problem set, and an informal comparison between data from the competition and the results presented here, indicate that it outperforms $\text{HSP}_a^*$ as well). In the non-temporal `promela` and `psr` domains, other optimal planners also outperform $\text{HSP}_a^*$ (competition results are presented by Hoffmann and Edelkamp, ?). CPT also uses the temporal $h^2$ heuristic, but searches a CSP formulation of partial order planning: heuristic estimates and the current cost bound are formulated as constraints, and cost bound violations are inferred by constraint propagation, avoiding the need to explicitly evaluate states and enabling earlier detection. The partial order branching and the use of efficient propa-

---

14. These "components" are sets of propositions with the property that exactly one proposition in the set is true in any reachable world state, *i.e.*, invariants. Thus, in any state exactly one (`at-segment aircraft ?segment`) is true for each `aircraft`, and so on (the case when the aircraft is airborne is special) gation are both important for the efficiency of the CPT planner, but it would probably also benefit from a more accurate heuristic. Thus, the search scheme of CPT and the idea of improving heuristics through search are complementary, and may be possible to combine.

## 5. Related Ideas

The idea of using search to derive or improve heuristics is not new. This section reviews a selection of related methods. With the exception of the discussion of pattern database heuristics, the focus in this section is on how these (or similar) methods can be adapted and exploited to improve relaxed search.

### 5.1 Search-Based Heuristics

Deriving heuristics by solving an abstracted, or relaxed, version of the search problem is not a new idea, and neither is the idea of using search to solved the abstract problem (see *e.g.* Gaschnig, 1979; Pearl, 1984; Prieditis, 1993).

A recent, and successful, variant on this theme is *pattern database heuristics* (Culberson & Schaeffer, 1996; Hernadvölgyi & Holte, 2000). These are defined by abstracting away part of the problem and solving only the part that remains (the "pattern"). The abstraction implicitly defines a projection from the problem search space into a smaller search space: optimal solution cost in this abstract space is a lower bound on optimal solution cost in the original search space, and by making this space small enough the optimal cost of every state in the abstract problem space can found by blind search, and stored in a table so that state evaluation can be done by a simple table lookup (hence the name pattern database). Heuristic estimates from multiple abstractions can be combined by taking their maximum (in some cases their sum; see Felner, Korf, & Hanan, 2004). Pattern database heuristics have been successfully applied to a number of standard search problems (Culberson & Schaeffer, 1996; Felner et al., 2004), and also to sequential STRIPS planning (Edelkamp, 2001). The idea of pattern databases may appear very similar to relaxed search (indeed, to the definition of the $h^m$ heuristics in general) in that the problem is "split" into simpler problems which are solved independently, and the solution costs for these used as a heuristic estimate for the complete problem. There is, however, a crucial difference, in that the $h^m$ relaxation performs this split *recursively*, to every state of size more than $m$, while the abstraction that defines the pattern in a pattern database heuristic is fixed (also, the abstraction that defines a PDB is in general a *projection*, *i.e.*, a many-to-one mapping, while the $h^m$ relaxation is an *embedding*, *i.e.*, each state in the original regression search space corresponds to a single state in the relaxed search space). Even if estimates from multiple pattern databases (abstractions) are combined, the combination of values (by maximizing or summing) occurs only at the root, *i.e.*, the state being evaluated, and not along the solution paths from this state in the different abstractions. This difference means that in some problems, the $h^m$ heuristic can be more accurate than any pattern database of reasonable size (Haslum, Bonet & Geffner, 2005, provide an example). On the other hand, the possibility of admissibly summing values from multiple pattern databases means that in some problems, a collection of additive pattern databases can form a heuristic more accurate than $h^m$, for any reasonable value of $m$ (again, Haslum, Bonet & Geffner, 2005, give an example).

In some ways, relaxed search has more in common with the idea of *pattern searches*, developed in the context of the Sokoban puzzle, which are more dynamic (Junghanns & Schaeffer, 2001). Like in a pattern database heuristic, a pattern search abstracts away part of the problem and solves the remaining (small) problem to obtain an improved lower bound, but the pattern (*i.e.*, the part of the problem that is kept by the abstraction) is selected whenever a particular state expansion ("move") is considered. Patterns that have been searched are stored, along with their updated cost, and taken into account in the heuristic evaluation (by maximization) of any new state that *contains* the same pattern, encountered by the search. The patterns explored by pattern searches are found through an incremental process: The first pattern consists of only the part of the problem ("stone") that is directly affected by the move under consideration. The next pattern extends the previous with stones that in the current state conflict with the solution found in the preceding pattern search, and this is repeated until no more conflicts are found.

As mentioned in the previous section, the high computational cost of the $m$-regression search is often due to the fact that AND-nodes have many successors, and most of the time it is not actually necessary to search them all for every AND-node: an alternative is to search only the most "promising" successors, where a promising successor is an OR-node whose cost is likely to be underestimated by the current heuristic, and therefore likely to increase when the node is expanded. Limiting the number of successors searched for every AND-node uniformly to at most $k$ results in an $(m, k)$-regression space, and a series of $(m, k)$-regression searches with increasing $m$ and $k$ can be organized in different ways: for example, the planner could perform $(m, 1)$-regression searches for $m = 3, \ldots$ until some suitable stopping condition is met, then $(m, 2)$-regression searches for $m = 3, \ldots, etc.$ This is very similar to iterative broadening search (Ginsberg & Harvey, 1992). An alternative is to limit the expansion of AND-nodes non-uniformly, for example, to search all successors satisfying some criterion for being promising, similar to the iterative widening strategy used in the context of game-tree search (Cazenave, 2001).

The conflict-directed strategy used to find patterns in pattern searches demonstrates a way to distinguish promising successors. Consider sequential planning where an AND-node is simply a set of more than $m$ (subgoal) atoms, and the successors are all size $m$ subsets of this set: subsets more likely to have a higher cost than the estimate given by $h^{m-1}$ can be identified by solving each size $m - 1$ subset and examining the solution for conflicts with remaining atoms in the state (a conflict being for example an action that deletes the atom, or actions incompatible with an action needed to establish the atom). In fact, this method has been used in the context of a different method for improving the $h^m$ heuristics through limited search (Haslum, 2004a). Some care must be taken to ensure that the searches needed to find the promising sets are not more expensive than searching every set, but if the $h^{m-1}$ value was also computed by relaxed search, the size $m - 1$ subsets (or at least some of them) have already been searched, and conflicts found during previous searches can be saved.

## 5.2 IDAO* vs. SCOUT's Test and Other Algorithms for AND/OR Graph Search

The SCOUT AND/OR tree search algorithm, developed mainly for searching game trees, tries to reduce the number of nodes evaluated by first testing for each node if it can affect the value of its parent before evaluating the node exactly (Pearl, 1984). The test is performed by a procedure, called simply "Test", which takes as arguments a node and a threshold value, and determines if the value of the node is greater (or equal) than the threshold, by recursively testing the node's successors (to a specified depth), but only until the inequality is proven. The procedure can easily be modified to return a greater value for the node when such a value is found, though this may still be less than the nodes actual value, and it has been shown that the Test procedure, enhanced with memory in the form of a transposition table, can be used iteratively to give an efficient algorithm that determines the exact value of a node (Plaat, Schaeffer, Pijls, & A., 1996).

The DFS subroutine of IDAO* is very similar to a depth-unbounded version of the Test procedure, and thus the IDAO* algorithm is similar to such an iterative application of Test. The main difference lies in that IDAO-DFS applies iterative deepening (by calling IDAO*) to the successors of AND-nodes, whereas Test calls itself recursively with the same cost bound. As a result, IDAO* finds the optimal cost of any solved OR-node, which Test does not (though the higher cost returned by (modified) Test when the cost of a node is proven to exceed the threshold is still a lower bound on the nodes optimal cost).

Recently, Bonet and Geffner (2004) presented a general depth-first search algorithm for AND/OR-graphs, called LDFS, which is also very similar to IDAO*. Like IDAO*, it finds the optimal cost of every solved node, and improved lower bounds on nodes that are explored but not solved. LDFS, however, stops searching the successors of an AND-node as soon as one of them is found to have a cost greater than the current estimate for that node, whereas IDAO* performs iterative deepening searches until the node is solved or shown to have a cost greater than the current estimated cost of the predecessor AND-node. Experiments with an Iterative Test algorithm for $m$-regression search have shown that it is not more efficient than IDAO*. An experimental comparison between IDAO* and the LDFS algorithm remains to be done.

IDAO*, Iterative Test and LDFS all perform top-down, depth-first iterative deepening searches, but none of these characteristics of the algorithms are essential for the their use in computing an improved heuristic. Any AND/OR search algorithm can be used to carry out the relaxed search, as long as it discovers the optimal cost (or a greater lower bound) of every expanded OR-node. For example, standard AO* (Nilsson, 1968) and the Generalized Djikstra algorithm by Martelli and Montanari (1973) both do this, and both offer a possibility of trading greater memory requirements for less search time (though the results of Bonet and Geffner, 2004, indicate that this may not be the case).

AND/OR search has been mostly investigated in the AI area of game playing. The AND/OR (or Min-Max) search spaces representing two-player games are somewhat different from the $m$-regression space: there is no concept of solution cost, other than the won/lost distinction, and for most games it is infeasible to search for a complete solution, rather the search aims to improve the accuracy of a heuristic estimate of the usefulness of a move or position in the game. Thus, game-tree searches are depth-bounded, rather than cost-bounded, and values at the leaf nodes of the tree are given by a static heuristic function. $m$-regression can be formulated in this way, by taking the sum of accumulated and estimated cost as the static heuristic function. A depth-bounded search with a standard game-tree search algorithm (Pearl, 1984; Plaat et al., 1996) can be used to improve the accuracy of the estimated cost of the root node. This, however, fails to achieve the main objective of relaxed search, which is to discover (and store) improved cost estimates for the size $m$ states encountered during the search, so this method would have to be used in a different way, *e.g.*, as a depth-bounded look-ahead to improve the accuracy of heuristic evaluations of states in the normal regression search. On the other hand, since the improved heuristic value is in this mode of use not stored but only used for the state from which the look-ahead search originates, it is not necessary to simplify the $m$-regression space, and a potentially more powerful relaxation can be used.

## 6. Conclusions (Reprise)

The two planners I entered in the 2004 International Planning Competition, TP4 and $\text{HSP}_a^*$, are very similar: the only difference is that $\text{HSP}_a^*$ invests some effort into computing a more accurate heuristic, through a series of searches in relaxed regression spaces (the $m$-regression spaces) which are derived from the same relaxation as the $h^2$ heuristic used by TP4. Indeed, the motivation for entering both planners was to make use of the competition as an experimental evaluation of the relaxed search technique, as well as comparing both planners to other state-of-the-art optimal temporal planners. For several reasons however, the competition results did not provide the complete picture of the relation between $\text{HSP}_a^*$ and TP4.

As demonstrated here, $\text{HSP}_a^*$ can produce better results than TP4 in some of the competition domains, though in no domain does $\text{HSP}_a^*$ completely dominate over TP4. It is mainly on problems that are hard, for both planners, that the heuristic improvement resulting from relaxed search yields an advantage. In some domains, the improvement over the $h^2$ heuristic is not enough to compensate for the time spent computing it. A more detailed analysis resulted in a (weak) characterization of the domains in which relaxed search can be expected to be cost effective: in such domains, expanding small states is computationally cheaper than expanding large states, and small states tend to have small successor states. In the domains where it is too expensive, on the other hand, the central problem is that the branching factor of $m$-regression is *higher* than that of normal regression search, so that searching in the relaxed space is computationally *more* expensive (quite contrary to the idea of obtaining a heuristic estimate by solving a "simplified" problem). Two measures, *viz.* the branching factor of OR-nodes in relaxed search relative to branching factor in the original search space and the relative size of the successors of OR-nodes in relaxed search, where found to be good indicators of how well $\text{HSP}_a^*$ performed, compared to TP4, in a given domain. In the experiments, these measures were taken by exploring the search spaces, but it may also be possible to estimate them (if not to calculate them exactly) from the domain description.

The analysis of the domains in which relaxed search fails to be effective also points out possibilities for improvement, and several ideas for potential improvements to the method can be found in related "incremental" search schemes in the literature. These include limiting search to a smaller fraction of the relaxed space, using conflicts to direct search to the states more likely to be have their heuristic values improved, and alternative search algorithms for AND/OR graphs. This is one direction for future developments.

Finally, results clearly demonstrate that although improving the heuristic improves performance in some domains, alone it is not enough to achieve good performance reliably across all the competition domains. CPT (Vidal & Geffner, 2004), the only other optimal temporal planner to participate in the competition, appears to yield better results in most competition domains (though no precise comparison has been made). CPT, like TP4, uses the temporal $h^2$ heuristic, but performs a non-directional search, and uses constraint representation and propagation techniques to infer cost bound violations. The inefficiency of directional search in temporal planning has been noted before: thus CPT and HSP$_a^*$ can be said to improve on two different weaknesses of TP4. Combining these improvements is another challenge for the future.

## Acknowledgments

Héctor Geffner has had an important part in the development of relaxed search since its inception and through numerous discussions. Jonas Kvarnström, Per Nyblom, JAIR associate editor Maria Fox and the anonymous reviewers all provided very helpful comments on drafts of the paper. All faults are of course mine.